\documentclass[letterpaper]{article} 
\usepackage{aaai25}  
\usepackage{times}  
\usepackage{helvet}  
\usepackage{courier}  
\usepackage[hyphens]{url}  
\usepackage{graphicx} 
\usepackage{natbib}  
\usepackage{caption} 
\usepackage{algorithm}
\usepackage{algorithmic}

\usepackage{newfloat}
\usepackage{listings}
\DeclareCaptionStyle{ruled}{labelfont=normalfont,labelsep=colon,strut=off} 
\floatstyle{ruled}
\newfloat{listing}{tb}{lst}{}
\floatname{listing}{Listing}
\usepackage{amsmath}
\usepackage{multirow}
\usepackage{booktabs}
\usepackage[table]{xcolor}

\title{MVReward: Better Aligning and Evaluating Multi-View \\Diffusion Models with Human Preferences}
\author{
    Weitao Wang\textsuperscript{\rm 1}\equalcontrib, Haoran Xu\textsuperscript{\rm 2}\equalcontrib, Yuxiao Yang\textsuperscript{\rm 1}, Zhifang Liu\textsuperscript{\rm 1}, Jun Meng\textsuperscript{\rm 2}\thanks{Corresponding Authors.}, Haoqian Wang\textsuperscript{\rm 1}\footnotemark[2]
}
\affiliations{
    \textsuperscript{\rm 1}Tsinghua Shenzhen International Graduate School, Tsinghua University\\
    \textsuperscript{\rm 2}Zhejiang University\\
}

\usepackage{bibentry}

\begin{document}

\maketitle

\begin{abstract}
Recent years have witnessed remarkable progress in 3D content generation. 
However, corresponding evaluation methods struggle to keep pace. Automatic approaches have proven challenging to align with human preferences, and the mixed comparison of text- and image-driven methods often leads to unfair evaluations.
In this paper, we present a comprehensive framework to better align and evaluate multi-view diffusion models with human preferences.
To begin with, we first collect and filter a standardized image prompt set from DALL·E and Objaverse, which we then use to generate multi-view assets with several multi-view diffusion models. Through a systematic ranking pipeline on these assets, we obtain a human annotation dataset with 16k expert pairwise comparisons and train a reward model, coined MVReward, to effectively encode human preferences.
With MVReward, image-driven 3D methods can be evaluated against each other in a more fair and transparent manner.
Building on this, we further propose Multi-View Preference Learning (MVP), a plug-and-play multi-view diffusion tuning strategy. Extensive experiments demonstrate that MVReward can serve as a reliable metric and MVP consistently enhances the alignment of multi-view diffusion models with human preferences.
\end{abstract}

%

\section{Introduction}

3D content generation is developing rapidly, thanks to the powerful generation ability of 2D diffusion and increasing efforts \cite{dreamfusion, magic3d} to lift its rich priors to 3D. Given a text or image prompt, current methods can generate 3D objects with detailed geometry and texture in seconds, with multi-view diffusion models playing an indispensable role. These models pave the way for fast and robust 3D generation by training view-aware diffusions to generate consistent images. 
As the 3D generation field gains more attention, its further development is highly anticipated.

\begin{figure}[t]
\centering
\includegraphics[width=0.44\textwidth]{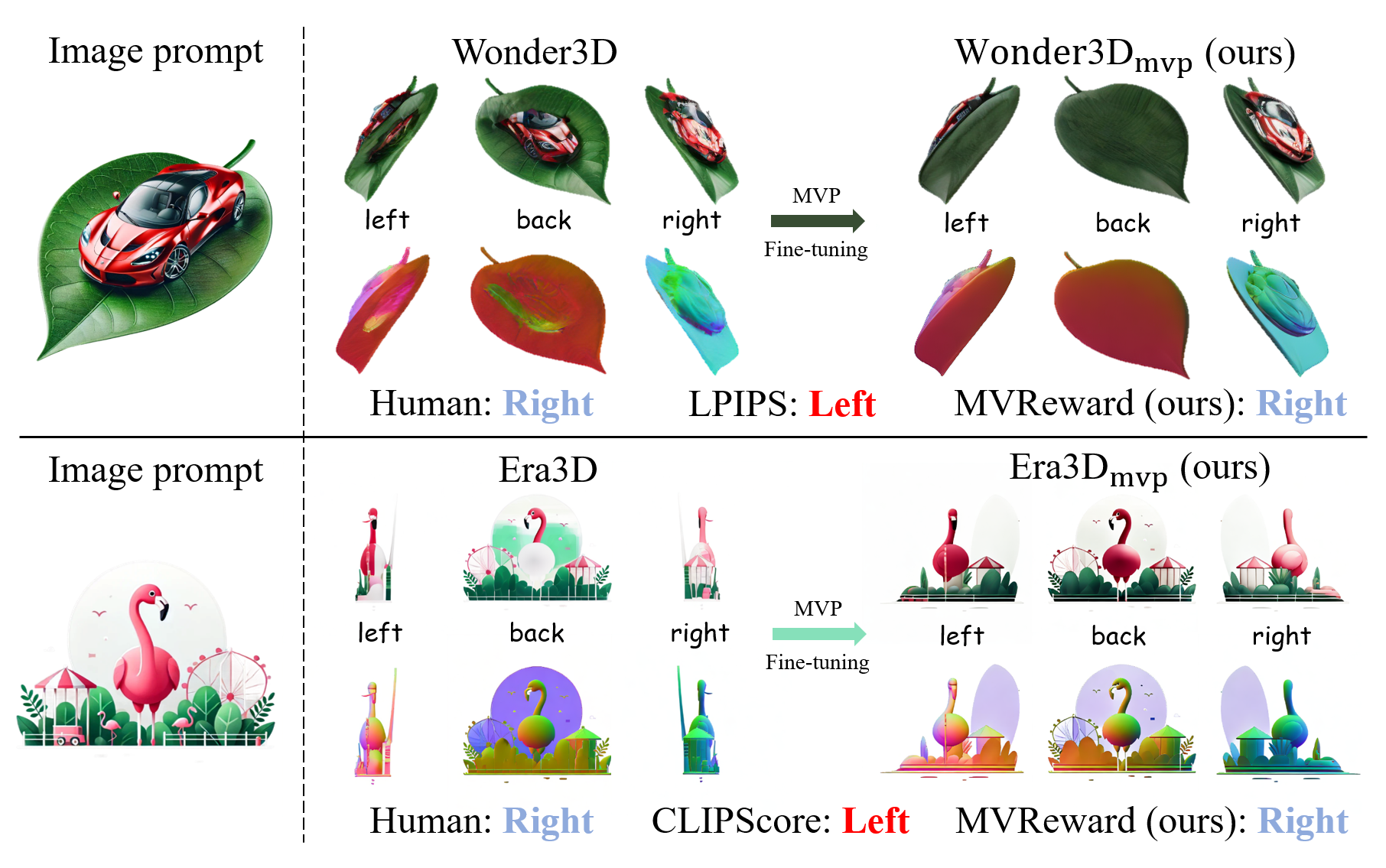} 
\caption{Automatic metrics often struggle to align with human preferences in evaluating image-to-3D tasks. Our MVReward model fills this gap and our MVP further enhances the alignment of existing multi-view diffusion models with human preferences.}
\label{first}
\end{figure}

Nonetheless, the lack of corresponding 3D evaluation methods is becoming a significant obstacle to progress in this area. As fully illustrated in previous research \cite{toward, gpteval}, existing automatic methods such as FID, LPIPS, CLIPScore, etc. often fail to align with human preferences, as shown in Figure \ref{first}. Furthermore, taking the widely-used GSO dataset \cite{gso} as an example, the test objects are not consistently unified across different methods, exacerbating this misalignment. Meanwhile, comparing generated content to ground truth (GT) data in an ill-posed task, is a one-sided approach that may stifle the model's creative potential. The everyday objects in the GSO dataset also lack adequate evaluation of those imaginative, unrealistic objects that are common in generative tasks. Considering the reasons above, most 3D generation methods rely heavily on qualitative analysis---comparing results through the subjective judgment of the viewer which may differ greatly across individuals. 

Several works \cite{imagereward, hps} attempt to apply RLHF (Reinforcement Learning from Human Feedback) methods to text-to-3D tasks to solve the problem of misalignment with human preferences. However, similar research is absent in the image-to-3D field, particularly for multi-view diffusion methods, and the evaluation biases brought by the GSO dataset have not been taken seriously.

\begin{figure*}[t]
\centering
\includegraphics[width=0.8\textwidth]{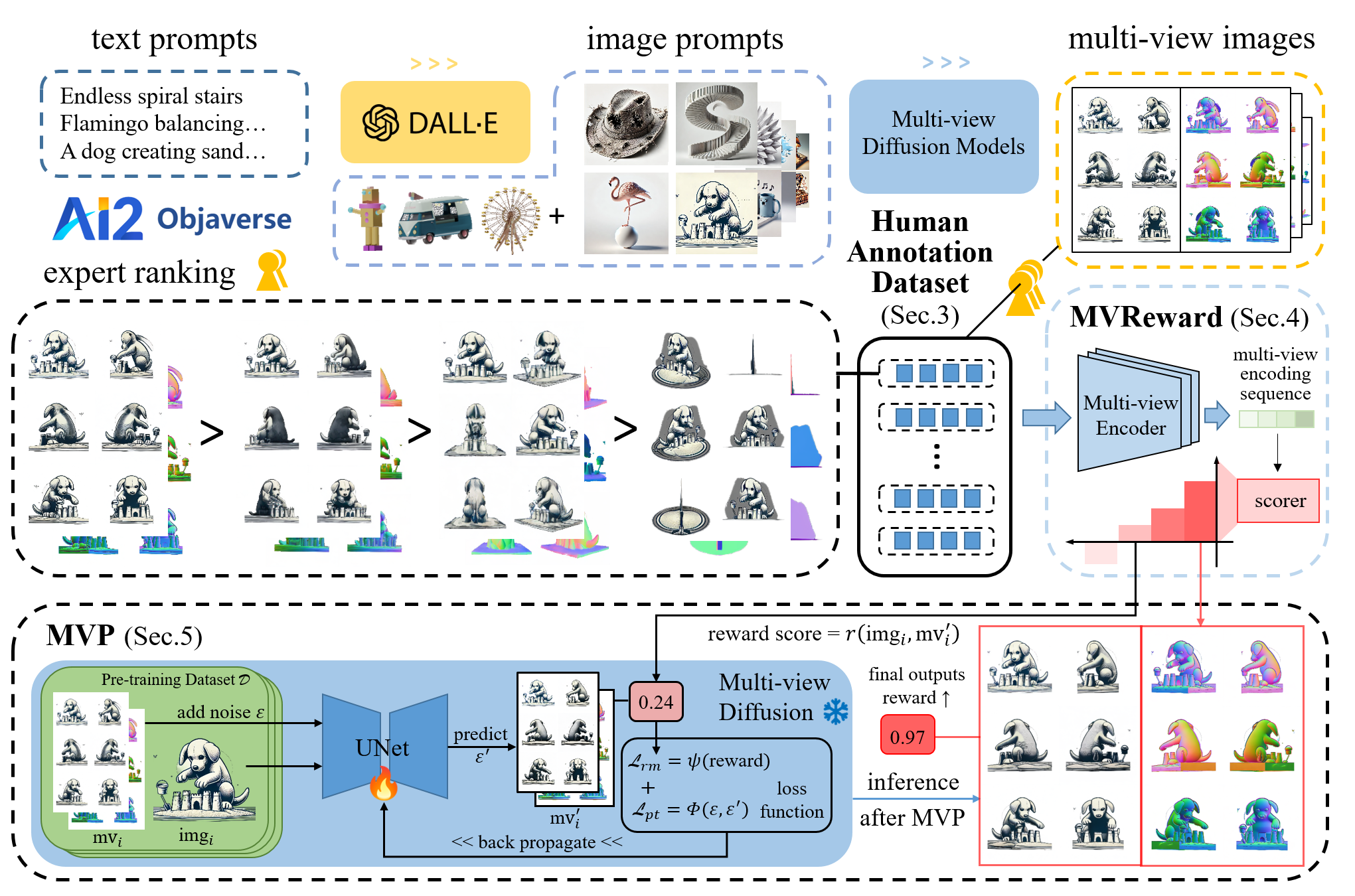} 
\caption{An overview of our whole framework. Our human annotation dataset is constructed by a text prompt $\Rightarrow$ image prompt $\Rightarrow$ multi-view images $\Rightarrow$ human annotation procedure (Sec.3). Then we train our MVReward model, which includes a multi-view encoder and a scorer to effectively encode human preferences and evaluate multi-view images (Sec.4). Finally we propose MVP to fine-tune multi-view diffusion models by combining pre-trained loss with our reward loss (Sec.5).}
\label{pipe}
\end{figure*}

On the other side, the mixed evaluation of text- and image-driven methods is also becoming a serious issue. Image-to-3D tasks have distinct application areas from text-to-3D, and images, as more fine-grained inputs, require higher quality than text. Some evaluation methods \cite{gpteval} simply use the text-to-image results of text prompts as inputs for image-driven methods like \cite{wonder3d} during 3D evaluation. Since image-driven methods typically remove the inputs' background for better generation results, we observe incomplete objects and missing scenes like the example in Figure \ref{text}, causing potential unfair comparisons between text- and image-driven methods.

In light of these challenges, in this paper we introduce a complete framework specially designed for multi-view diffusion based methods, providing a normalized image-to-3D evaluation environment and better alignment with human preferences. We begin by generating and filtering a standardized image prompt set from DALL·E \cite{dalle} and Objaverse \cite{objaverse}, ensuring the object(s) in each image are fully visible with well-designed geometry and texture. Then we select four multi-view diffusion methods to generate RGB and normal multi-view assets using image prompts. After annotating and collecting 16k expert pairwise comparisons of these multi-view assets as a human annotation dataset, we train the MVReward model on it. We further propose MVP, a plug-and-play tuning strategy to align multi-view diffusion models with human preferences.

Our main contributions are as follows:

\begin{itemize}
    \item We systematically analyze the chaos and challenges in the evaluation of image-driven 3D methods, and based on this, create a comprehensive pipeline that includes filtering a standardized image prompt set, generating high-quality multi-view assets, and collecting 16k expert pairwise comparisons as a human annotation dataset.
    \item We train MVReward, the first general-purpose human preference reward model for multi-view diffusion, which can serve as a reliable image-to-3D metric.
    \item We present MVP, a plug-and-play multi-view diffusion tuning strategy, consistently enhancing the alignment of multi-view diffusion models with human preferences.
\end{itemize}

\section{Related Work}

\subsection{3D Generation with Multi-view Diffusion}

Current 3D generation techniques can be broadly classified into three categories: distillation based methods, multi-view based methods and feed-forward methods. Distillation based methods \cite{sjc, prolificdreamer, fantasia3d} aim to leverage the strong priors of 2D diffusion models to generate consistent 3D representations like NeRF by designing a Score Distillation Sampling (SDS) loss, but suffer from low efficiency and multi-face problem. Multi-view based methods \cite{zero123, syncdreamer} alleviate this issue by directly fine-tuning 2D diffusion models to generate multi-view images without relying on 3D representation and time-consuming per-shape optimization. Feed-forward methods \cite{instant3d, lrm, lgm, crm, grm, instantmesh} seek to generate 3D shapes end-to-end in seconds, by integrating off-the-shelf multi-view diffusions with sparse-view large reconstruction models. It is evident that nowadays multi-view diffusion plays a significant role in 3D generation. Zero123 \cite{zero123} first trains a diffusion model conditioned on the reference image and camera viewpoint, pioneering the multi-view generation. More advancements \cite{v3d, sv3d} are emerging after it and the potential of the multi-view diffusion in 3D generation continues to be explored.


\subsection{Learning from Human Feedback}

Pre-trained generative models often exhibit misalignment with human intent due to the limited size and inherent biases of the training data. Reinforcement Learning from Human Feedback (RLHF) is a commonly used method to address this deviation. Natural language processing (NLP) methods \cite{drl, rlhp} first align the language model with human preference via training a reward model (RM). InstructGPT \cite{instructgpt} further applies RLHF to GPT-3, improving its performance significantly in multi-task NLP. ImageReward \cite{imagereward} and Pick-a-pic \cite{pickscore} expand the application scope of RLHF to text-to-image generation by collecting human annotation image datasets and training RMs on them. RLHF methods in 3D generation \cite{dreamreward} are also emerging, but research on aligning multi-view images with human preference remains blank and our MVReward fills it.

\subsection{3D Generation Evaluation}

Evaluating 3D generation has long been a challenging task, requiring rich 3D priors and a deep understanding of physical properties. Existing 3D methods typically rely on automatic metrics along with user studies for comparison. The GSO \cite{gso} and Omni3D \cite{omni3d} datasets are widely used to measure the distance between the generated novel views (or 3D shapes) and the reference ones using PSNR, SSIM, LPIPS (or Chamfer Distance, Volume IOU). In the absence of a reference set, multimodal pre-trained embeddings such as CLIP \cite{clip} and BLIP \cite{blip} may also be utilized to calculate similarities between different views. 
Recent work \cite{gpteval, t3bench} develop several GPT4V \cite{gpt4} intervention procedures to leverage its strong priors in 3D perception. Concurrently, RLHF methods \cite{dreamreward, hfdream} seek to train reward models for text-to-3D evaluation. Different from the above, our method proposes a standardized evaluation process for image-driven 3D methods, especially the multi-view based ones.

\section{Human Annotation Dataset}
Our whole framework is clearly overviewed in Figure \ref{pipe}. In this section, we elaborate on the construction of our human annotation dataset, which is prepared for our MVReward training and future research in human preference learning. 

\subsection{Text Prompts}
Text prompts remain essential in image-to-3D tasks by controlling the details of image prompts using text-to-image models. In practice, without advanced knowledge of the potential image input, we often simulate the image prompt distribution by crafting suitable text prompts. 

\begin{figure}[t]
\centering
\includegraphics[width=0.44\textwidth]{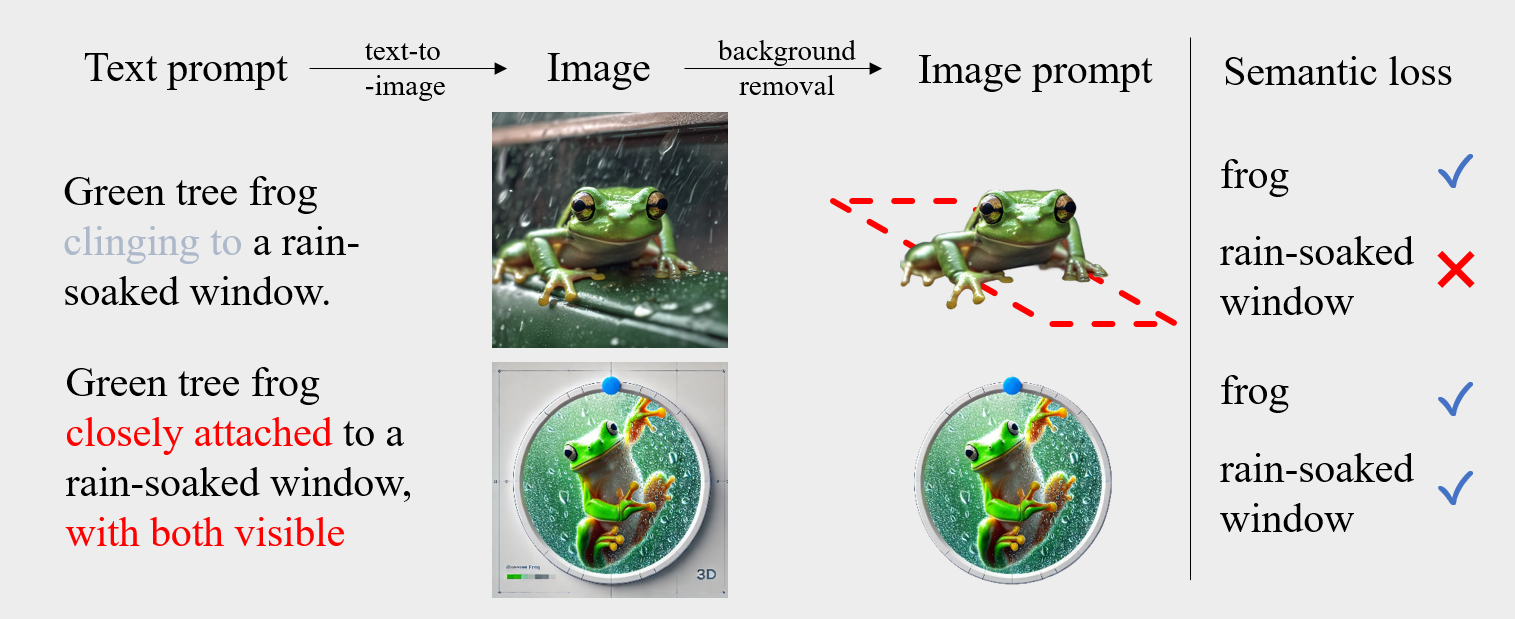} 
\caption{Example of our text prompt enhancements to prevent potential semantic loss brought by the background removal of image-to-3D methods.}
\label{text}
\end{figure}

Our text prompt set is mainly derived from text-to-3D evaluation methods \cite{t3bench, gpteval}, with 
necessary enhancements due to the inherent biases in text- and image-to-3D tasks. For instance, the background of the input image is typically removed by image-to-3D methods to ensure the object-centric generation quality. Therefore, for text prompts with background descriptions, we delete or edit them by adding strong conjunctions like {\em 'closely attached to ... with both visible'} instead of {\em 'clinging to'} to prevent semantic loss of the image prompts during background removal (see Figure \ref{text}). 
In addition, to emphasize the geometry and texture that are of interest in image-to-3D tasks, text prompts are filtered and more generated with increasing complexity and creativity in mind, leveraging GPT4V (details in Appendix B.1).

After a clustering algorithm \cite{tsne} standardizing and unifying the distribution of text prompts, we finally obtain 600 high-quality text prompts for image prompt generation.

\subsection{Image prompts}
Image prompts are crucial for both generation and evaluation in the image-to-3D task. As discussed in Section 1, the absence of a standardized image prompt set hampers fair evaluation, further limiting 3D generation progress. 
To solve this issue, we generate image prompts for two key purposes: constructing a human annotation dataset for training an efficient and robust reward model, and providing a standardized set for fair comparisons across image-to-3D methods.

Our image prompt set is expected to exhibit increasing complexity and creativity in geometry and texture, owing to our complicated construction of the text prompt set. For example, {\em a gift wrapped with mysterious symbols} is a common object with simple geometry but complex texture, whereas {\em a swan with feathers resembling origami folds} is a creative object with intricate geometry but plain texture. 
We use the high-quality text-to-image model DALL·E \cite{dalle} to generate samples and carefully select those suitable for multi-view tasks (see Appendix B.2). Figure \ref{prompt} demonstrates some examples from our image prompt distribution. For each text prompt, we select two distinct samples: one for the reward model training and one for the standardized set.

Meanwhile, we include the same number of non-generative, real-world objects from Objaverse \cite{objaverse} to increase the diversity of the dataset. We filter the front render view of each Objaverse object as its image prompt. 
In the end, we get an image dataset with 1200 images as input to multi-view diffusion models and a standardized image prompt set with 600 images for fair comparison.


\begin{figure}[t]
\centering
\includegraphics[width=0.42\textwidth]{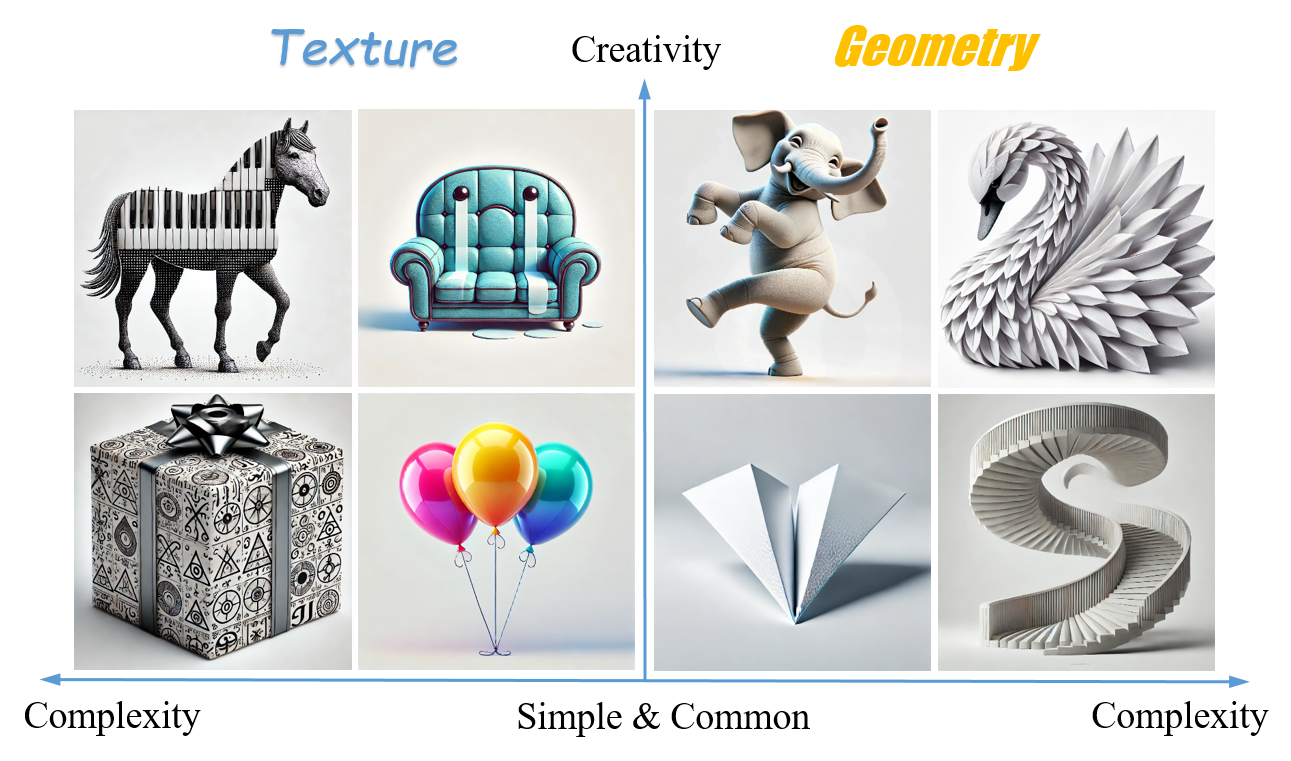} 
\caption{Examples from the image prompt distribution. Images closer to the edges represent objects with more complex and creative geometry or texture, while those near the center are mainly simple and common.}
\label{prompt}
\end{figure}

\subsection{Human annotation pipeline}
To grant our reward model with sufficient understanding of multi-view images quality, we construct a human-annotated multi-view images dataset. Firstly, we choose four high-quality multi-view diffusion methods: Wonder3D \cite{wonder3d}, Zero123++ \cite{zero123++}, Envision3D \cite{envision3d} and Era3D \cite{era3d} following the rule that each method can generate at least six views of RGB and normal images from an image prompt. 
Although camera systems and poses vary slightly across methods, our hypothesis, supported by empirical results, is that the reward model can evaluate multi-view images under different camera settings just like humans. For an image prompt, we generate two multi-view assets per method using different inference steps (20, 40), resulting in 10200 multi-view assets (122k images) in total, including rendered views from Objaverse.

The next step is to annotate these assets to capture human preferences. We develop a detailed annotator guideline in advance, including task objectives, evaluation criteria, etc. (see Appendix C). 20 annotators are carefully gathered, all with at least a bachelor's degree and an interest in 3D generation. The annotators rank 4-5 multi-view assets from the same image prompt, allowing ties if quality is close. Each annotator is allowed to rank up to 400 asset-lists to minimize fatigue and personal bias. We also provide our own annotations as {\em researcher annotations} for double checking.

The above annotation pipeline yields 7692 valid rankings. For conflicting annotations, we use the borda count to determine final rankings. At last, we obtain the rankings for 1200 multi-view asset-lists, resulting in 16k valid comparison pairs for the reward model training (except for ties).

\section{MVReward: Encoding Human Preferences}

We decompose the problem of evaluating multi-view images with human preferences into three sub-problems: \\
- Evaluating the quality of the generated images themselves.\\
- Assessing the degree of alignment between the generated images and the input image prompt.
\\
- Calculating the spatial and semantic consistency across the generated images.

To this end, we propose MVReward, which incorporates two core modules: a multi-view encoder for sub-problems 1 and 2 by encoding the full representation of a 3D object, and a scorer for sub-problem 3 by capturing and evaluating the inner connections between different views.

\begin{figure}[t]
\centering
\includegraphics[width=0.42\textwidth]{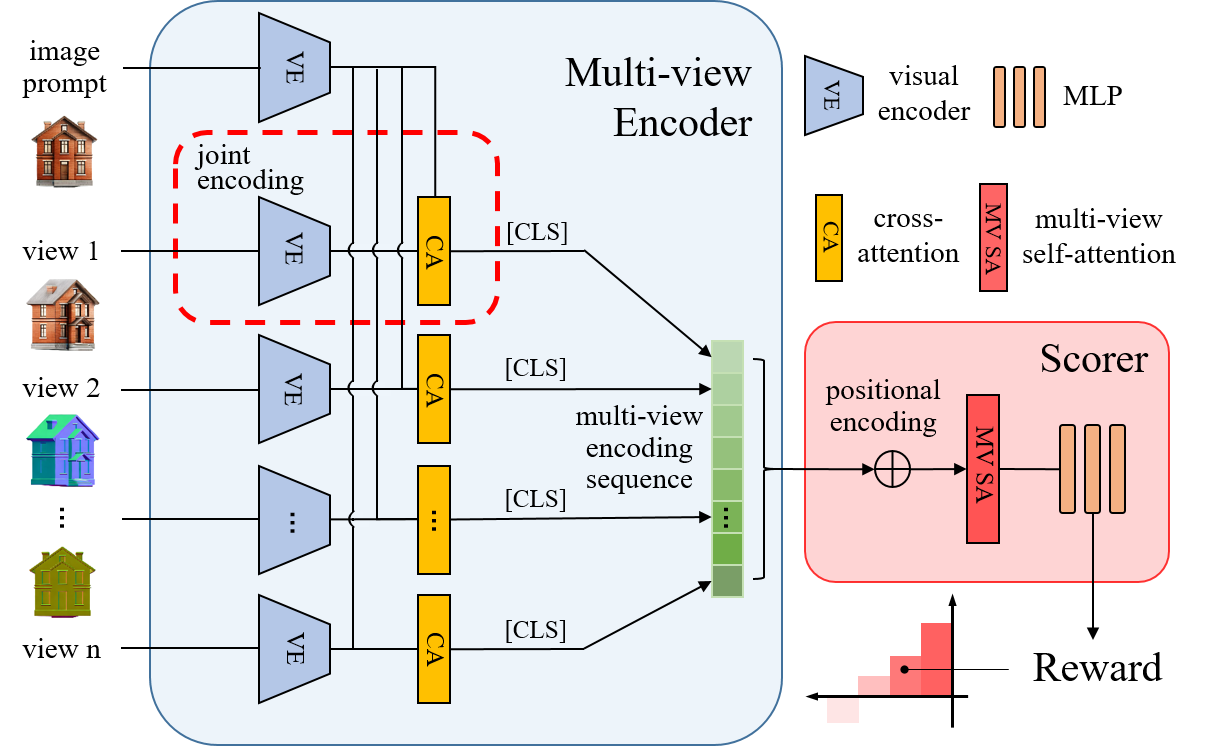} 
\caption{MVReward architecture with a multi-view encoder and a scorer to encode and predict human preferences.}
\label{mvreward}
\end{figure}

\subsection{Architecture Design}

Figure \ref{mvreward} illustrates the architecture of MVReward. We adapt BLIP \cite{blip} as the backbone for our multi-view encoder. The vanilla BLIP model encodes both image and text inputs into a unified representation space, capturing similarities between the two modalities. We extend this approach to connect the input image prompt with the generated views, treating them as distinct modalities.

The multi-view encoder module comprises multiple visual encoders, each corresponding to a different view, but sharing the same parameters. These encoders independently extract features from the image prompt and the generated views. We then apply a cross-attention layer to integrate these features, resulting in a joint encoding for each input-generated view pair.

For each view pair encoding, a special [CLS] token is placed at the beginning of the joint encoding to represent the global feature. These tokens are extracted to form a multi-view encoding sequence, which serves as a comprehensive representation of the views from a 3D asset (e.g., a set of 6$\times$[RGB + normal] images generated by a multi-view diffusion model paired with their corresponding input views worth a sequence with 12 tokens).

The scorer module processes this multi-view encoding sequence, using positional encoding to distinguish between different views and domains (such as RGB and normal).  The sequence then undergoes a multi-view self-attention layer to capture correlations between views. Each token from the processed sequence will be concatenated and passed through an MLP which outputs a scalar value indicating how well the generated multi-view images align with human preferences.

\subsection{MVReward Training}
We train MVReward using a cross-entropy loss function, similar to those used in prior work \cite{learn} for RM training. For each comparison pair sampled from the human annotation dataset, associated with the same image prompt $I$ and two sequences of generated views $s_w = (v_w^{(0)}, v_w^{(1)}, …, v_w^{(n)})$, $s_l = (v_l^{(0)}, v_l^{(1)}, …, v_l^{(n)})$, if the sequence $s_w = (v_w^{(0)}, v_w^{(1)}, …, v_w^{(n)})$ is better aligned with $I$, the loss function will be formulated as:
\begin{equation}
    \mathcal{L}_{\gamma} = -\mathbf{E}_{(I,s_w,s_l) \sim \mathcal{H}}[\text{log}(\sigma(r_{\gamma}(I,s_w)-r_{\gamma}(I,s_l)]
\end{equation}

where $\gamma$ denotes the learnable parameters of MVReward, and $\mathcal{H}$ represents the training human annotation dataset.

We further enhance the model's domain-aware evaluation by distinguishing between features from different modalities (e.g., RGB and normal) and providing targeted feedback. Positional encoding is used to correlate the generated view modality with its position in the input sequence. We also introduce a set of modality-reversed negative samples, where the RGB and normal modalities are swapped. This contrastive approach enables the model to better understand the impact of positional changes across modalities, improving its modality-aware capabilities.


\section{MVP: Multi-View Preference Learning}


Building on MVReward's ability to encode human preferences for multi-view images, we further propose a plug-and-play tuning strategy to enhance the performance of multi-view diffusion models in aligning with human preferences.

\subsection{Preliminaries} 

\subsubsection{Multi-view diffusion training.} 
Most multi-view diffusion models are fine-tuned from 2D diffusion models (DM) which lack awareness of image-view correlations. With the advent of large-scale 3D datasets \cite{objaverse, omni3d}, 3D priors have become enough for DMs to perceive the underlying correlations across different views. Existing multi-view DMs typically leverage the multi-view attention mechanisms to enhance consistency among views. Many methods like Wonder3D \cite{wonder3d}, push a step forward by adding cross-domain attention to extend the generation to the normal domain, enabling the model to generate both color images and normal maps simultaneously. During the training process, the same noise is added to different views and domains (in Wonder3D, it's 12 images) of the same object. The DM UNet then predicts the noise across views, maintaining the L2 loss between predicted and added noise used in 2D diffusion.

\subsubsection{Reward feedback learning (ReFL).} 
ReFL, as proposed by ImageReward \cite{imagereward}, fine-tunes a text-to-image latent diffusion model (LDM) by predicting $x_t \rightarrow x^{\prime}_0$ during the denoising steps. Empirical experiments show that the reward scores for generation $x^{\prime}_0$ between 75\%-99\% of denoising steps are reliable for feedback learning. The mid-step t is randomly selected instead of using the last step to avoid an unstable fine-tuning process. After re-weighting the reward loss with the pre-training loss, ReFL demonstrates its effectiveness through impressive visual results.

\begin{algorithm}[tb]
\caption{Multi-View Preference Learning (MVP) for Multi-View DMs}
\label{alg:algorithm}
\textbf{Pre-training Dataset: } Image-MV dataset $\mathcal{D}$~=~\{(img$_1$, mv$_1$), ..., (img$_n$, mv$_n$)\}, \textbf{where} mv$_i$= 6$\times$[RGB+normal]$_i$\\
\textbf{Input:} Multi-view DM UNet with pre-trained parameters $w_0$, reward model $r$, multi-view DM pre-training loss function $\phi$, reward-to-loss map function $\psi$, re-weighting scale $\lambda$\\
\textbf{Initialization:} The noise scheduler $N_s$ and VAE of multi-view DM and time steps $\mathcal{T}$
\begin{algorithmic}[1] 
\FOR{(img$_i$, mv$_i$) $\in$ $\mathcal{D}$}
\STATE $\varepsilon \sim \mathcal{N}(0,I)$ ~~~~~~~~~~~~// sample noise to be added to mv
\STATE latent $\leftarrow$ mv$_i$ + $\varepsilon$ ~~~~~~~~~~~~~~~~~~~~~~~~~~~~~~~~~~[with $N_s$, $\mathcal{T}$]
\STATE embeds $\leftarrow$  pre-training process
\STATE $\varepsilon^{\prime}$ $\leftarrow$ UNet$_{w_i}$(latent, embeds, $\mathcal{T}$) ~// predict the noise
\STATE $\mathcal{L}_{pt}$ $\leftarrow$ $\phi$~($\varepsilon, \varepsilon^{\prime}$) ~~~~~~~~~~~~~~~~~~~~~~~~~~~~~~~~~// pre-training loss
\STATE mv$_i^{\prime}$ $\leftarrow$ latent - $\varepsilon^{\prime}$ ~~~~~~~~~~~~~~~~~~~~~~~~~~[with $N_s$, $\mathcal{T}$, VAE]
\STATE $\mathcal{L}_{rm}$ $\leftarrow$ $\psi$~($r$~(img$_i$, mv$_i^{\prime}$)) ~~~~~~~~~~~~~~~~~~~~~~// reward loss
\STATE $w_{i+1}$ $\leftarrow$ $w_{i}$ ~~~~~~~~~~~~// update UNet~[with $\lambda$$\mathcal{L}_{pt}$+$\mathcal{L}_{rm}$] 

\ENDFOR
\end{algorithmic}
\end{algorithm}

\begin{table*}[htbp]
\centering
\begin{tabular}{lcccccccccc}
\toprule
{\multirow{2}{*}{Metrics \& Model}} & \multicolumn{2}{c}{Human Eval.} & \multicolumn{2}{c}{MVReward} & \multicolumn{2}{c}{GSO Dataset} & \multicolumn{2}{c}{CLIP} & \multicolumn{2}{c}{BLIP} \\
\multicolumn{1}{c}{} & Rank & Favor (\%) & Rank & Reward & Rank & LPIPS $\downarrow$ & Rank & Score & Rank & Score \\ \midrule
Envision3D & 7 & 0.06 & 7 & 0.205 & 3 & 0.130 & 6 & 0.774 & 4 & -0.253 \\ 
SyncDreamer & 6 & 0.10 & 6 & 0.690 & 7 & 0.146 & 7 & 0.759 & 2 & 0.057 \\ 
\multirow{3}{*}{\begin{tabular}[c]{@{}l@{}}Wonder3D\\   ~~- Wonder3D$_\text{pt}$\\  ~~- Wonder3D$_\text{mvp}$ (ours)\end{tabular}} & 5 & 9.33 & 5 & 1.016 & 5 & 0.141 & 5 & 0.785 & 7 & -0.262 \\
 & - & - & - & 1.018 & - & 0.141 & - & 0.790 & - & -0.264 \\
 & \cellcolor{gray!20}4 &  \cellcolor{gray!20}13.7 & \cellcolor{gray!20}4 & \cellcolor{gray!20}1.389 & \cellcolor{gray!20}6 & \cellcolor{gray!20}0.142 & \cellcolor{gray!20}4 & \cellcolor{gray!20}0.796 & \cellcolor{gray!20}5 & \cellcolor{gray!20}-0.258 \\  
Zero123++ & 3 & 21.0 & 3 & 1.475 & 4 & 0.133 & 1 & \textbf{0.822} & 1 & \textbf{0.886} \\
\multirow{3}{*}{\begin{tabular}[c]{@{}l@{}}Era3D\\   ~~- Era3D$_\text{pt}$\\  ~~- Era3D$_\text{mvp}$ (ours)\end{tabular}} & 2 & 25.3 & 2 & 1.506 & 2 & 0.126 & 3 & 0.801 & 6 & -0.259 \\
 & - & - & - & 1.503 & - & 0.127 & - & 0.802 & - & -0.259 \\
 & \cellcolor{gray!20}1 & \cellcolor{gray!20}\textbf{29.0} & \cellcolor{gray!20}1 & \cellcolor{gray!20}\textbf{1.676} & \cellcolor{gray!20}1 & \cellcolor{gray!20}\textbf{0.124} & \cellcolor{gray!20}2 & \cellcolor{gray!20}0.810 & \cellcolor{gray!20}3 & \cellcolor{gray!20}-0.248 \\ 
\bottomrule
\multicolumn{1}{c}{Spearman to Human Eval.} & \multicolumn{2}{c}{-}  & \multicolumn{2}{c}{\cellcolor{gray!20}\textbf{1.00}} & \multicolumn{2}{c}{0.61} & \multicolumn{2}{c}{0.86} & \multicolumn{2}{c}{0.04}  \\ \hline
\end{tabular}
\caption{Quantitative comparisons of MVReward and MVP. Our MVReward outperforms all other metrics in aligning with human preferences, and our MVP consistently improves the performance of fine-tuned methods. The \textbf{bold} is the best results under certain metrics and the \colorbox{gray!20}{gray} represents our results.}
\end{table*}

\begin{figure*}[htbp]
\centering
\includegraphics[width=0.85\textwidth]{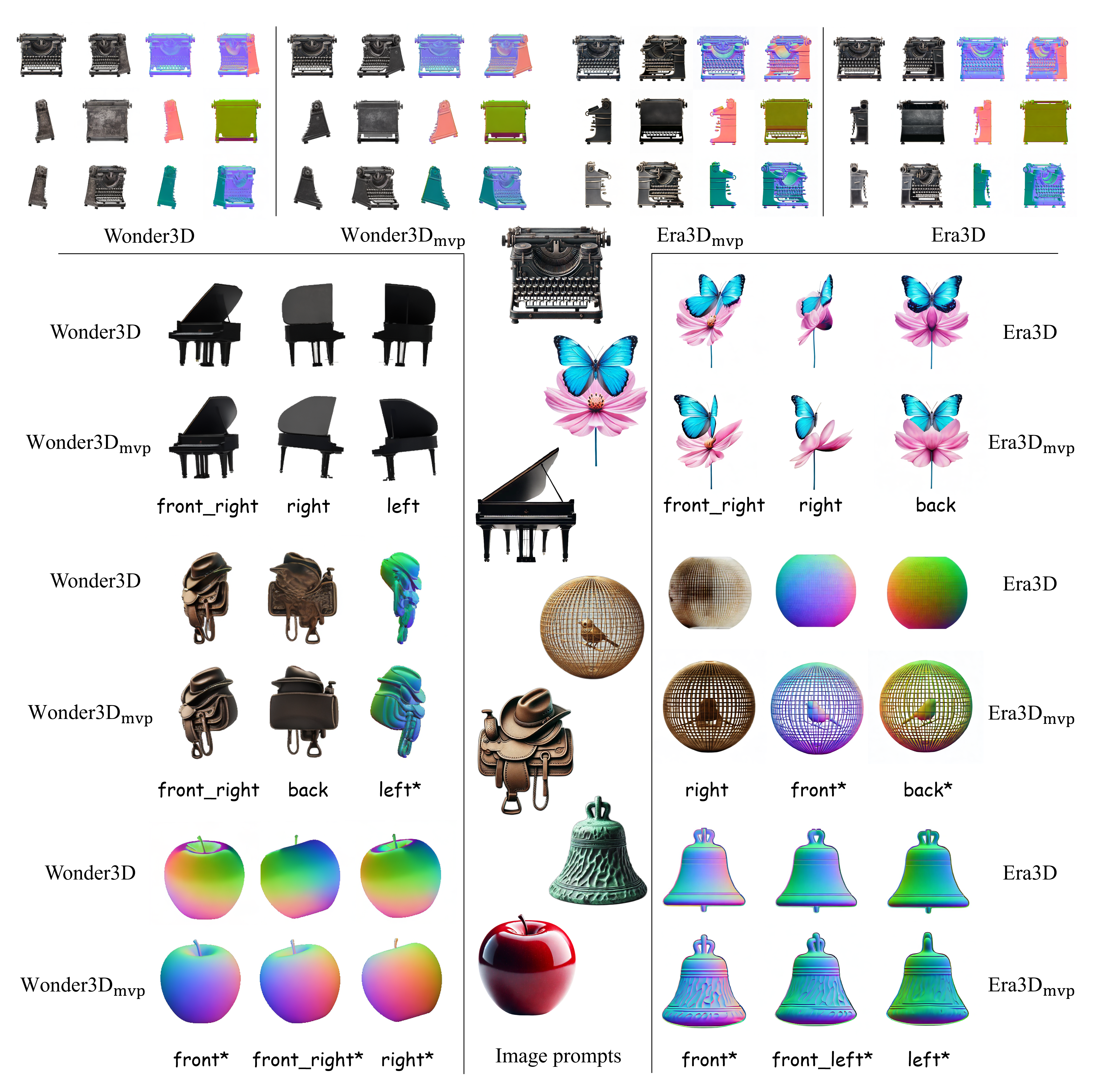} 
\caption{Qualitative comparisons between the original and the MVP fine-tuned multi-view diffusion models.}
\label{quanti}
\end{figure*}
\subsection{Multi-View Preference Learning}

Inspired by ReFL, we present Multi-View Preference Learning (MVP) to fine-tune multi-view diffusion models using MVReward. Following the standard multi-view diffusion pre-training, we add the same random noise to all ground-truth multi-view images and predict it with the pre-trained UNet, retaining the L2 loss between predicted and added noise. We then estimate the original latent through one-step sampling and decode them into images using the pre-trained VAE. The images are then fed into MVReward with their image prompt to obtain a reward score, which is converted into a reward loss through a reward-to-loss function map $\psi$. This reward loss is combined with the pre-training loss during backpropagation.

In practice, we observe significant generation deviations from the image prompt when simply adding reward loss to the pre-trained loss. Thus similar to ReFL, we introduce a large re-weighting scale 
$\lambda$ to the pre-training loss, ensuring strong pre-training constraints. The complete process is summarized in Algorithm 1, with the final loss function:
\begin{equation}
\begin{aligned}
\centering
    \mathcal{L} ~~~= ~~&\lambda \mathcal{L}_{pt} + \mathcal{L}_{rm} \\
    \mathcal{L}_{rm} = ~~&\mathbf{E}_{\text{img}_i \sim \mathcal{D}}(\phi(r(\text{img}_i, g_{\theta}(\text{img}_i)))) \\
    \mathcal{L}_{pt} ~= ~~&\mathbf{E}_{(\text{img}_i,\text{mv}_i)\sim \mathcal{D}}(E_{\epsilon(\text{mv}_i),\text{img}_i,\epsilon \sim N(0,1),t}\\&~~~~~~~~~~~~~~[\|\epsilon - \epsilon_\theta(z_t,t,\tau_{\theta}(\text{img}_i)) \|_2^2])
\end{aligned}
\end{equation}

where $\theta$ is the multi-view DM parameters, and $g_{\theta}(\text{img}_i)$ denotes the predicted \text{mv}$_i$ from the multi-view DM with $\theta$ and image prompt. $\mathcal{L}_{pt}$ depends on the specific multi-view diffusion model to be fine-tuned. In our case it's the loss function derived from \cite{ldm}.

\section{Experiment}

In this section, we present the quantitative and qualitative comparisons of MVReward and MVP against other existing methods to demonstrate their superiority. Limited by space, additional experimental results can be found in Appendix D.

\subsection{MVReward: Predicting human preferences and evaluating multi-view images}

\subsubsection{Dataset \& Training settings.} MVReward is trained on the human annotation dataset constructed in Section 3, with 1200 multi-view images asset-lists and 16k comparison pairs in total. The training, validation and test datasets are split according to an 8:1:1 ratio. The multi-view encoder is initialized using the pre-trained BLIP/VIT-B checkpoint. To prevent overfitting and ensure training stability, we fixed 50\% of its parameters. Optimal performance is achieved with a batch size of 96 in total, an initial learning rate of 4e-5 using cosine annealing, on 4 NVIDIA Quadro RTX 8000.


\subsubsection{Quantitative results.} We conduct a user study to evaluate MVReward's ability in predicting human preferences. We collect multi-view assets from 7 methods on our standardized image prompt set and ask 6 new annotators to choose their favorite. For methods incapable of generating normal maps, we use GeoWizard \cite{geowizard} for prediction. Table 1 shows that our MVReward outperforms the commonly used GSO Dataset (LPIPS), CLIP Score, and BLIP Score in aligning with human preferences. 

\subsubsection{Win rates.} 
\begin{figure}[t]
\centering
\includegraphics[width=0.4\textwidth]{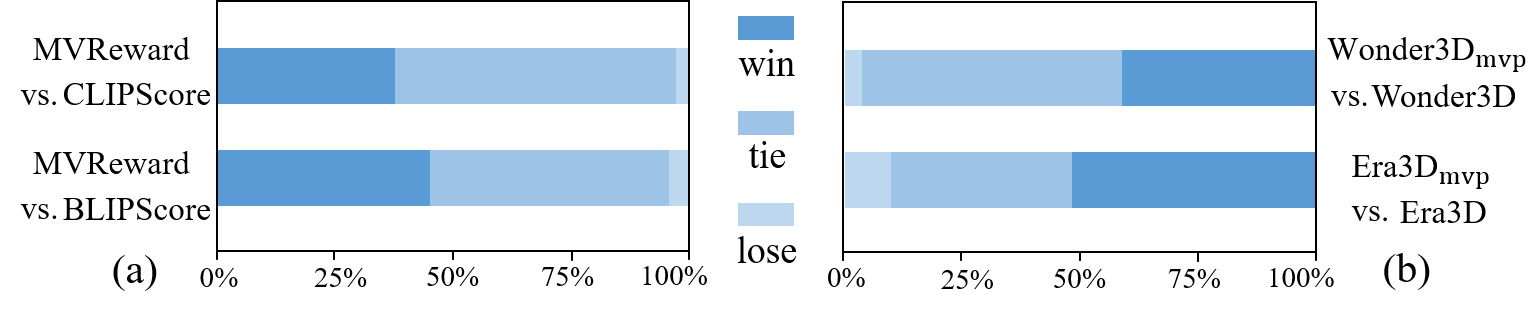} 
\caption{Win rates of MVReward and MVP. Our methods remain unbeaten versus their opponents most of the time.}
\label{wtl}
\end{figure}

We also calculate the win rates of MVReward versus other metrics in Figure \ref{wtl}(a). For each pairwise comparison, a metric wins if its preference aligns with human's while the opponent does not. A tie is recorded if two metrics have the same preference, which may occur frequently. Despite this, our MVReward still maintains a high win rate, demonstrating its strong alignment with human preferences.

\subsubsection{Ablation study.}
We perform ablation studies on the encoder backbone, multi-view self-attention, and negative samples to assess their effects on MVReward. Table 2 indicates that substituting either results in reduced accuracy.

\subsection{MVP: Aligning multi-view diffusion models with human preferences} 

\subsubsection{Baseline \& Training settings.} 
We select Wonder3D as our fine-tuning baseline since it is the only one of the four multi-view diffusion methods we used that has a publicly accessible training code. However, we also provide an unofficial fine-tuned version of Era3D to verify that MVP is a plug-and-play strategy. The model parameters are fixed except for the designated trainable modules within the UNet.
Both models are fine-tuned in half-precision on 8 NVIDIA Quadro RTX 8000, with a batch size of 128 in total and a learning rate of 5e-6 with warm-up.

\subsubsection{Quantitative results.} For a fair comparison, we continue training the two pre-trained models with only pre-trained loss for the same iterations as fine-tuning process, namely Wonder3D$_{\text{pt}}$ and Era3D$_{\text{pt}}$ in Table 1. 
Since their results are nearly identical to the original versions, we exclude them from the user study and rankings, reporting only their metric scores. Our fine-tuned models Wonder3D$_{\text{mvp}}$ and Era3D$_{\text{mvp}}$ outperform both the original and pt-only versions across all the metrics, demonstrating the effectiveness of our MVP.

\subsubsection{Qualitative results.} Figure 6 presents several representative visual examples, highlighting the geometry and texture improvements brought by MVP (* means the normal map).

\subsubsection{Win Rates.} Figure \ref{wtl}(b) illustrates the win rates of our fine-tuned models versus their original version during user study. Though tie occurs at times for simple objects, MVP consistently improves the generated multi-view images quality.

\subsubsection{Ablation study.} 
The comparison between the fine-tuned and pt-only version is an ablation study on the reward loss and we further develop it in Figure 8, illustrating that both the pre-trained loss and reward loss are essential in MVP.

\begin{table}[t]
\centering
\begin{tabular}{ccccc}
 & w/o BLIP & \begin{tabular}[c]{@{}c@{}}w/o MV SA\end{tabular} & \begin{tabular}[c]{@{}c@{}}w/o Neg.\end{tabular} & Full (ours) \\ \midrule
\begin{tabular}[c]{@{}c@{}} Acc.\end{tabular} & 81.7 & 81.5 & 74.8 & \textbf{83.1}\\
\end{tabular}
\caption{Ablation study of MVReward on backbone, multi-view self-attention (MV SA) and negative samples (Neg.), where w/o BLIP means changing the backbone to CLIP and Acc. denotes the evaluation accuracy of the reward model.}
\label{rewardabl}
\end{table}

\begin{figure}[t]
\centering
\includegraphics[width=0.48\textwidth]{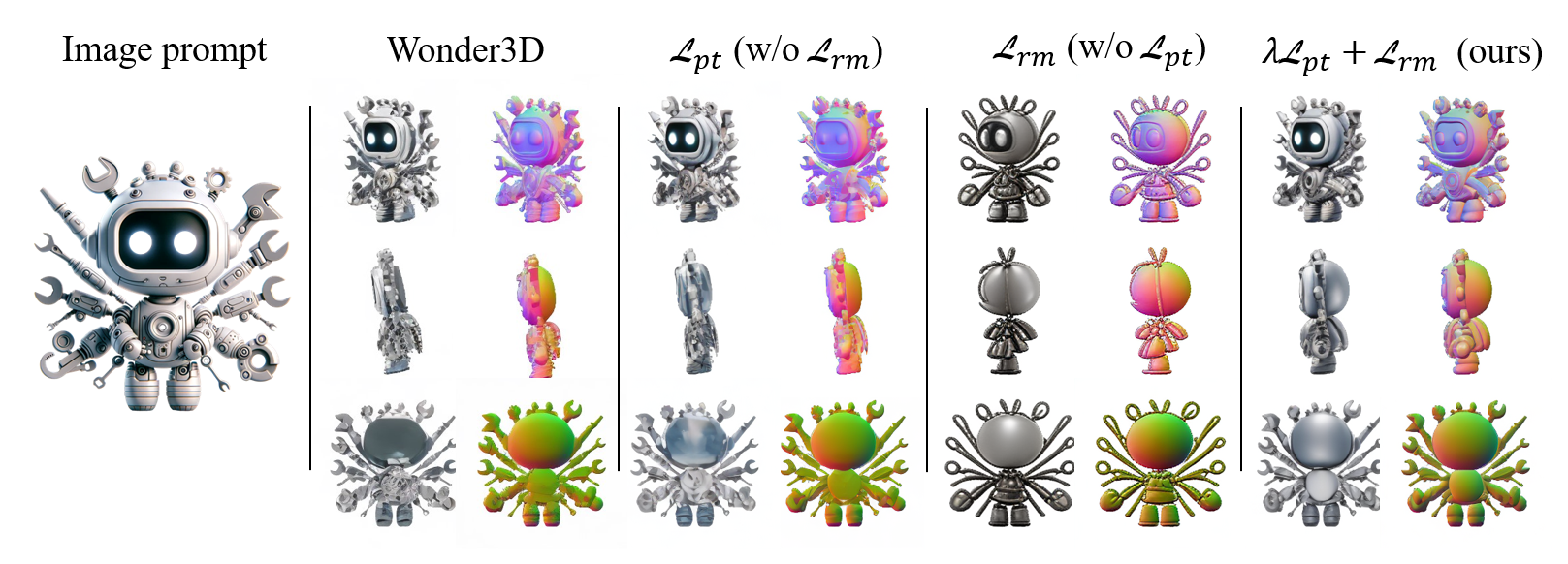} 
\caption{Further ablation experiment on the pre-trained loss and reward loss.}
\label{ablation}
\end{figure}

\section{Conclusion}

In this paper, we address the challenges in aligning and evaluating multi-view diffusion models with human preferences. We construct a human annotation dataset and a standardized image prompt set through a comprehensive pipeline. We then train MVReward, an effective reward model that can serve as a reliable image-to-3D metric. We also introduce MVP as a plug-and-play tuning strategy for multi-view diffusion models to enhance their performance in human preference alignment. Experimental results support our contributions in providing a fair evaluation method and an efficient improving approach for multi-view diffusion models.

\subsubsection{Limitations.} Although fruitful, we admit that our reward model requires more training data for better evaluation. Additionally, focusing on multi-view images instead of direct 3D representations (i.e. mesh) limits the scope of our approach. We will endeavor to tackle these issues in the future and look forward to more potential insights.

\bibliography{aaai25}

~\\~\\~\\~\\~\\~\\~\\~\\~\\~\\~\\~\\~\\~\\~\\~\\~\\~\\~\\~\\~\\~\\~\\~\\~\\~\\~\\~\\~\\~\\~\\~\\~\\~\\~\\~\\~\\~\\~\\~\\~\\~\\~\\~\\~\\~\\~\\~\\~\\~\\~\\~\\
\textbf{Reproducibility Checklist}\\
\textbf{Question1:} Includes a conceptual outline and/or pseudocode description of AI methods introduced? \\
Answer: Yes. \\
\textbf{Question2:} Clearly delineates statements that are opinions, hypothesis, and speculation from objective facts and results? \\
Answer: Yes. \\
\textbf{Question3:} Provides well marked pedagogical references for less-familiar readers to gain background necessary to replicate the paper? \\
Answer: Yes. \\
\textbf{Question4:} Does this paper make theoretical contributions? \\
Answer: No. \\
\textbf{Question5:} Does this paper rely on one or more datasets? \\
Answer: Yes. \\
\textbf{Question6:} A motivation is given for why the experiments are conducted on the selected datasets? \\
Answer: Yes. \\
\textbf{Question7:} All novel datasets introduced in this paper are included in a data appendix? \\
Answer: Yes. \\
\textbf{Question8:} All novel datasets introduced in this paper will be made publicly available upon publication of the paper with a license that allows free usage for research purposes? \\
Answer: Yes. \\
\textbf{Question9:} All datasets drawn from the existing literature (potentially including authors’ own previously published work) are accompanied by appropriate citations? \\
Answer: Yes. \\
\textbf{Question10:} All datasets drawn from the existing literature (potentially including authors’ own previously published work) are publicly available? \\
Answer: Yes. \\
\textbf{Question11:} All datasets that are not publicly available are described in detail, with explanation why publicly available alternatives are not scientifically satisficing? \\
Answer: NA. \\
\textbf{Question12:} Does this paper include computational experiments? \\
Answer: Yes. \\
\textbf{Question13:} Any code required for pre-processing data is included in the appendix? \\
Answer: Yes. \\
\textbf{Question14:} All source code required for conducting and analyzing the experiments is included in a code appendix? \\
Answer: Yes. \\
\textbf{Question15:} All source code required for conducting and analyzing the experiments will be made publicly available upon publication of the paper with a license that allows free usage for research purposes? \\
Answer: Yes. \\
\textbf{Question16:} All source code implementing new methods have comments detailing the implementation, with references to the paper where each step comes from ? \\
Answer: Yes. \\
\textbf{Question17:} If an algorithm depends on randomness, then the method used for setting seeds is described in a way sufficient to allow replication of results? \\
Answer: NA. \\
\textbf{Question18:} This paper specifies the computing infrastructure used for running experiments (hardware and software), including GPU/CPU models; amount of memory; operating system; names and versions of relevant software libraries and frameworks? \\
Answer: Yes. \\
\textbf{Question19:} This paper formally describes evaluation metrics used and explains the motivation for choosing these metrics? \\
Answer: Yes. \\
\textbf{Question20:} This paper states the number of algorithm runs used to compute each reported result? \\
Answer: Yes. \\
\textbf{Question21:} Analysis of experiments goes beyond single-dimensional summaries of performance (e.g., average; median) to include measures of variation, confidence, or other distributional information? \\
Answer: Yes. \\
\textbf{Question22:} The significance of any improvement or decrease in performance is judged using appropriate statistical tests (e.g., Wilcoxon signed-rank)? \\
Answer: Yes. \\
\textbf{Question23:} This paper lists all final (hyper-)parameters used for each model/algorithm in the paper’s experiments? \\
Answer: Yes. \\
\textbf{Question24:} This paper states the number and range of values tried per (hyper-) parameter during development of the paper, along with the criterion used for selecting the final parameter setting? \\
Answer: Yes. \\

\textbf{The detailed relevant information for Questions 13, 14, 18, 20, 21, 22, and 24 can be seen in the supplementary materials, and the other questions are all fully explained in our main paper.}

\end{document}